\documentclass[conference]{IEEEtran}
\IEEEoverridecommandlockouts
\usepackage{cite}
\usepackage{amsmath,amssymb,amsfonts}
\usepackage{algorithmic}
\usepackage{adjustbox}
\usepackage{graphicx}
\usepackage{textcomp}
\usepackage{xcolor}
\usepackage{booktabs}
\usepackage{multirow}
\usepackage{float}
\usepackage{hyperref}
\usepackage{url}
\usepackage{tabularx}
\usepackage{subcaption}
\usepackage{enumitem}
\usepackage{braket}
\usepackage{cite}  
\usepackage{blindtext}

\usepackage{geometry}
\usepackage{array}
\newcolumntype{L}[1]{>{\raggedright\let\newline\\\arraybackslash\hspace{0pt}}m{#1}}
\newcolumntype{C}[1]{>{\centering\let\newline\\\arraybackslash\hspace{0pt}}m{#1}}
\newcolumntype{R}[1]{>{\raggedleft\let\newline\\\arraybackslash\hspace{0pt}}m{#1}}

\def\BibTeX{{\rm B\kern-.05em{\sc i\kern-.025em b}\kern-.08em
    T\kern-.1667em\lower.7ex\hbox{E}\kern-.125emX}}

\begin{document}

\title{Quantum Machine Learning in Transportation: A Case Study of Pedestrian Stress Modelling\\
{\small IEEE Intelligent Transportation Systems Conference, 2025, Gold Coast, Australia}\\
	{\small Funding source: Natural Sciences and Engineering Research Council of Canada and Canada Research Chair program}
}

\author{\IEEEauthorblockN{Bara Rababah and Bilal Farooq}
\IEEEauthorblockA{Laboratory of Innovations in Transportation (LiTrans)\\
Toronto Metropolitan University, Toronto, Canada\\
brababah@torontomu.ca; bilal.farooq@torontomu.ca}
}

\maketitle

\begin{abstract}
Quantum computing has opened new opportunities to tackle complex machine learning tasks, for instance, high-dimensional data representations commonly required in intelligent transportation systems. We explore quantum machine learning to model complex skin conductance response (SCR) events that reflect pedestrian stress in a virtual reality road crossing experiment. For this purpose, Quantum Support Vector Machine (QSVM) with an eight-qubit ZZ feature map and a Quantum Neural Network (QNN) using a Tree Tensor Network ansatz and an eight-qubit ZZ feature map, were developed on Pennylane. The dataset consists of SCR measurements along with features such as the response amplitude and elapsed time, which have been categorized into amplitude-based classes. The QSVM achieved good training accuracy, but had an overfitting problem, showing a low test accuracy of 45\% and therefore impacting the reliability of the classification model. The QNN model reached a higher test accuracy of 55\%, making it a better classification model than the QSVM and the classic versions. 

\end{abstract}


\section{Introduction}
Machine learning techniques have become essential when analyzing complex datasets in intelligent transportation systems (ITS), where data relationships, such as those seen in crossing-related pedestrian stress experiments, require capturing higher-order correlations and precise detection and classification. In recent years, quantum machine learning has emerged as a novel approach that uses quantum computing to enhance the capabilities of classical algorithms \cite{shi2006}. Quantum models can encode data into a high-dimensional Hilbert space of quantum states and make use of quantum properties (superposition and entanglement) for complex computational needs. This approach could allow for more expressive representations than classical models, by effectively using an exponentially large feature space that is too complex for conventional methods \cite{biamonte2017}. In classification tasks, quantum support vector machines (QSVMs) and quantum neural networks (QNNs) have been introduced as quantum versions of classical SVMs and neural networks, with the potential of quantum-enhanced feature mapping and more efficient use of model parameters. 

This study explores the potential of quantum machine learning models, such as QSVMs and QNNs, in ITS with a real-world pedestrian stress dataset. Skin-conductance response (SCR), also called electrodermal activity, tracks changes in the skin's electrical conductance caused by sweat-gland activity. Because these fluctuations reflect sympathetic-nervous-system arousal, SCR, typically recorded with galvanic-skin-response (GSR) sensors, is widely used as a physiological indicator of stress or emotional arousal \cite{NAZEMI2025104952}. Important insights can be taken from the peaks in stress, allowing for the assessment of both cognitive and physiological factors that shape pedestrian experiences at road crossings.


We developed two quantum models on Pennylane quantum circuit emulator \cite{bergholm2018pennylane}, a QSVM utilizing a ZZ feature map for quantum kernel estimation, and a QNN using a variational circuit based on a Tree Tensor Network architecture. ZZ-circuit allows for pair-wise interactions, a medium level of complexity in terms of quantum circuit, and entanglement of quantum bits (qubits), making it highly suitable for complex datasets in ITS. The variational approach allows for fine-tuning of the quantum circuits using parameters, making them further suitable for ITS applications. We compare the performance of QSVM and QNN against a classical SVM (with RBF and linear kernels) and a classical deep neural network on the same classification task. \textcolor{black}{These models were selected because their structure is more compatible with quantum circuits. Not all machine-learning algorithms can be efficiently encoded into quantum-circuit encoding, so by focusing on SVMs and deep neural networks, we use a balanced comparison between classical and quantum formulations.}

\textcolor{black}{This study addresses key gaps in prior quantum machine learning  research in transportation, where most work is limited where it comes to real-world data, consistent baselines, and feature map evaluations. This study presents a direct comparison of QSVM and QNN against classical models on a real pedestrian stress dataset using identical inputs and preprocessing. By testing three quantum feature maps and selecting the ZZ map based on performance and entanglement.}

The goals of this paper are to (1) explore the implementation details of quantum machine learning models for transportation application, including the quantum feature mapping and circuit design, (2) discuss the suitability of theoretical frameworks from quantum machine learning, and (3) evaluate and compare the models using accuracy, precision, and recall. By examining these models on an identical SCR classification problem (exactly the same dataset, with the same task), we aim to highlight the current strengths and limitations of quantum machine learning concepts compared to those of the well-established classical approaches.

\textcolor{black}{\section{Background}}

\textcolor{black}{The ITS applications of Quantum Computing (QC) primarily include solving combinatorial optimization problems (e.g., traveller scales person problem) using Ising models \cite{zhuang2024quantum}. Quantum machine learning is a growing field of research that offers higher efficiency, accuracy, and speedup for classification and prediction problems. It is slowly gaining momentum in terms of adoption in ITS. Meghanat et al. \cite{10931853} used QNN to classify shadow regions in an image for safety applications in automated vehicles, reporting a faster detection than the classic counterparts. Hybrid QNN model has been used to develop traffic forecasting, showing a comparative performance with the state-of-the-art \cite{schetakis2025}. QML has also been used to find optimal routing solutions for small city networks \cite{mohanty2024solving} and in emergency situations \cite{haboury2023}. Another interesting application of QML is in the prediction of flight delays \cite{pophale2023performance}. Zhuang et al. \cite{zhuang2024quantum} developed a detailed survey on the use of QC in ITS. It points to the potential of achieving exponential speedup by reducing processing time from $O(n)$ to $O(log(n))$. Havlíček et al.\ \cite{Havlicek2019} demonstrated that entangling feature maps in quantum kernel classifiers capture complex, non-linear correlations in physiological signals that classical kernels struggle to model. Moreover, Goto et al. \cite{Goto2021} demonstrated that quantum feature spaces possess universal approximation properties for continuous functions, indicating that QML models may generalize more effectively on small, high-dimensional SCR datasets used in pedestrian stress experiments.}

\textcolor{black}{While in the recent literature, there have been a few cases where QML has been applied in traffic forecasting, delay prediction, and routing, to the best of our knowledge, their adoption in terms of pedestrian behaviour prediction, especially their neurophysiological state while crossing, has not been explored. Furthermore, a systematic comparison of multiple classic and quantum algorithms on a specialized ITS dataset also remains unexplored.}

\section{Dataset Description}

\begin{figure}[h]
\centering
\includegraphics[width=0.85\linewidth]{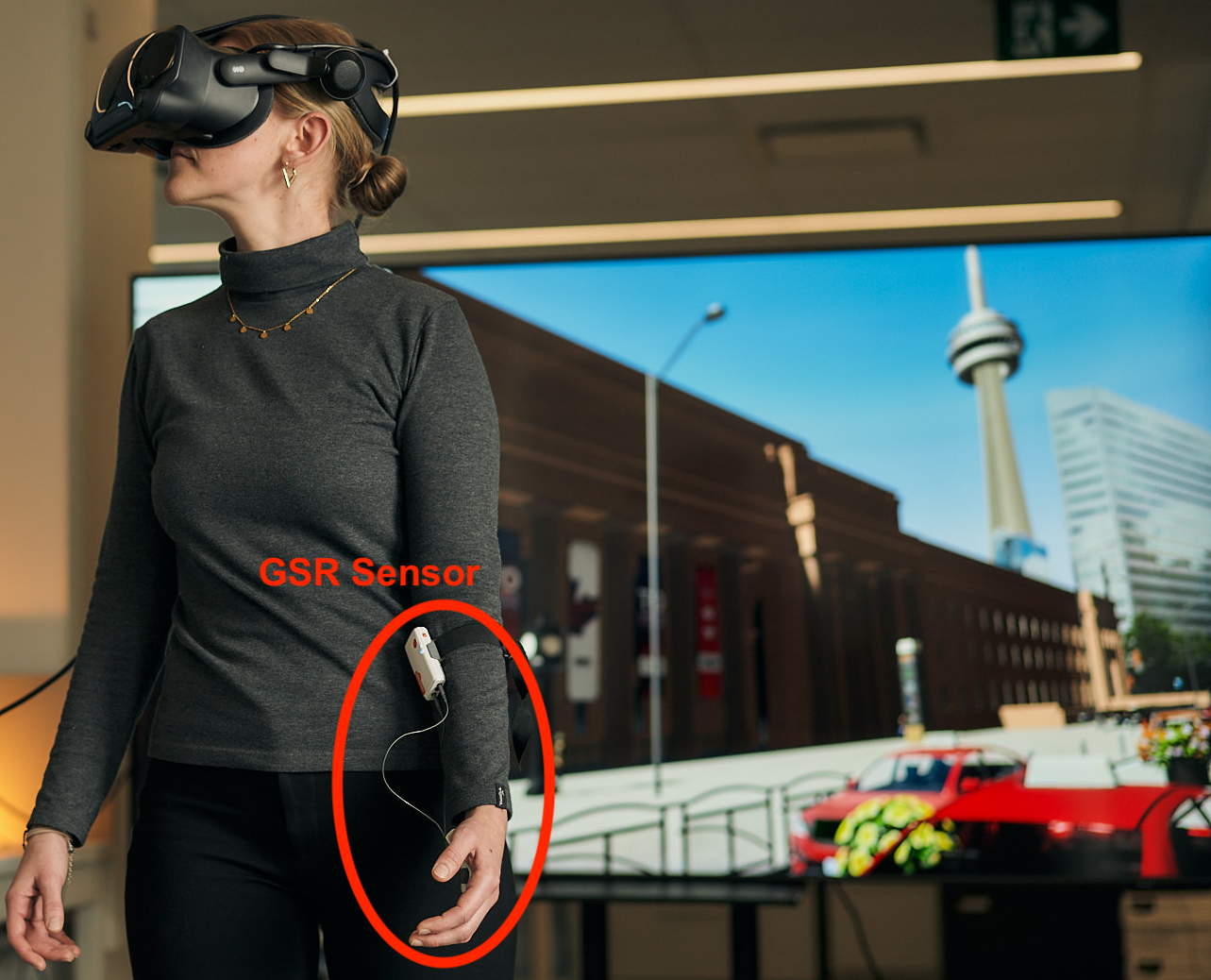}
\caption{VR environment \& GSR sensor \textcolor{black}{(demonstration only)}}
\label{fig:vrex}
\end{figure}

The dataset was originally collected in  Nazemi et al. \cite{NAZEMI2025104952} and contains stress measurements, in the form of Skin Conductance Response (SCR) values, collected from multiple participants in a VR based road-crossing experiment. \textcolor{black}{The data collection campaign was approved by the university ethics board with the identification number REB 2017-169.} The SCR values were collected from participants using GSR sensor (Fig. \ref{fig:vrex}). Each row in the dataset represents a single SCR event (a phasic increase in skin conductance) characterized by several features. Key features include the elapsed time in the session when the SCR occurred, the amplitude of the SCR (the magnitude of conductance change), the SCR (Skin Conductance Response value), and the count of SCR events detected up to that point in the session. The target variable is an amplitude class label that categorizes the SCR magnitude into discrete ranges to help identify the level of stress experienced.

The SCR amplitude is measured in micro Siemens units, and grouped into four classes, each having its own amplitude range:  $0.1 \leq \text{SCR} < 0.4$), class 1 (moderate: $0.4 \leq \text{SCR} < 0.7$), class 2 (high: $0.7 \leq \text{SCR} < 1.0$), and class 3 (very high: $\text{SCR} \geq 1.0$).
Each data sample is made up of four features: \texttt{elapsed\_time}, \texttt{scr\_amplitude}, \texttt{scr}, and \texttt{detected\_scr\_number}. The target variable is \texttt{amp\_class}. Table \ref{tab:sample_data} shows a sample of three data instances to illustrate the data format; the classes are listed based on correlation with the SCR level detected at that point.

\begin{table}[h]
\caption{Example data instances from the SCR dataset.}
\label{tab:sample_data}
\centering
\begin{adjustbox}{max width=\linewidth}
\begin{tabular}{C{1.6cm}C{1.8cm}C{0.7cm}cc}
\toprule
\textbf{elapsed\_time (s)} & \textbf{scr\_amplitude (µS)} & \textbf{scr (µS)} & \textbf{detected\_scr\_number} & \textbf{amp\_class} \\
\midrule
12.12 & 0.1059 & 1.74 & 1 & 0 \\
60.98 & 0.3558 & 1.81 & 2 & 0 \\
158.47 & 1.2000 & 2.50 & 5 & 3 \\
\dots & \dots & \dots & \dots & \dots \\
\bottomrule
\end{tabular}
\end{adjustbox}
\end{table}

\section{Methods and Models}

\begin{figure*}[htbp]
  \centering
  
  \begin{subfigure}[b]{\textwidth}
    \centering
    \includegraphics[width=0.7\linewidth]{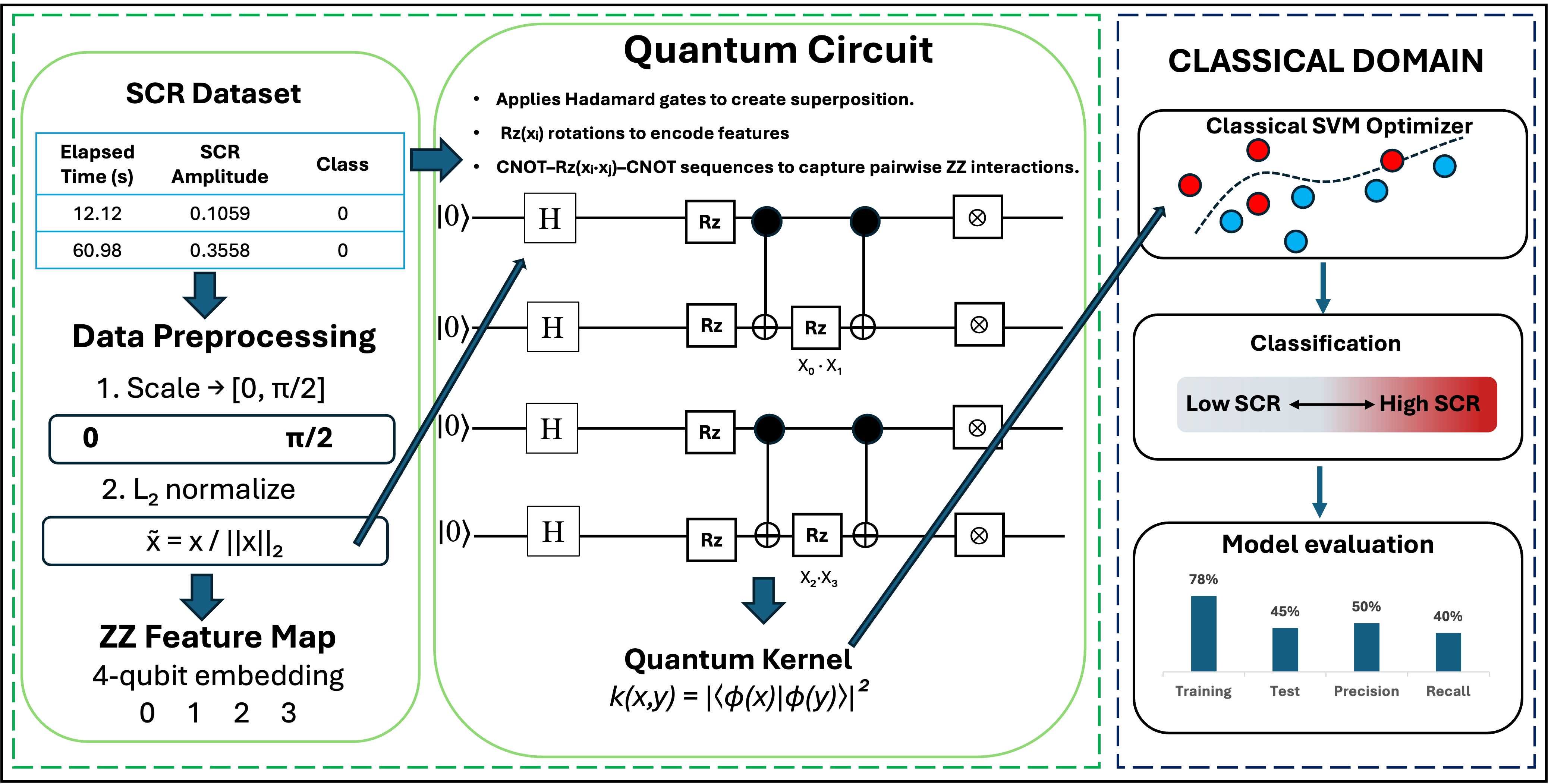}
    \caption{Quantum Support Vector Machine}
    \label{fma}
  \end{subfigure}
  \hfill
  \begin{subfigure}[b]{\textwidth}
    \centering
    \includegraphics[width=0.7\linewidth]{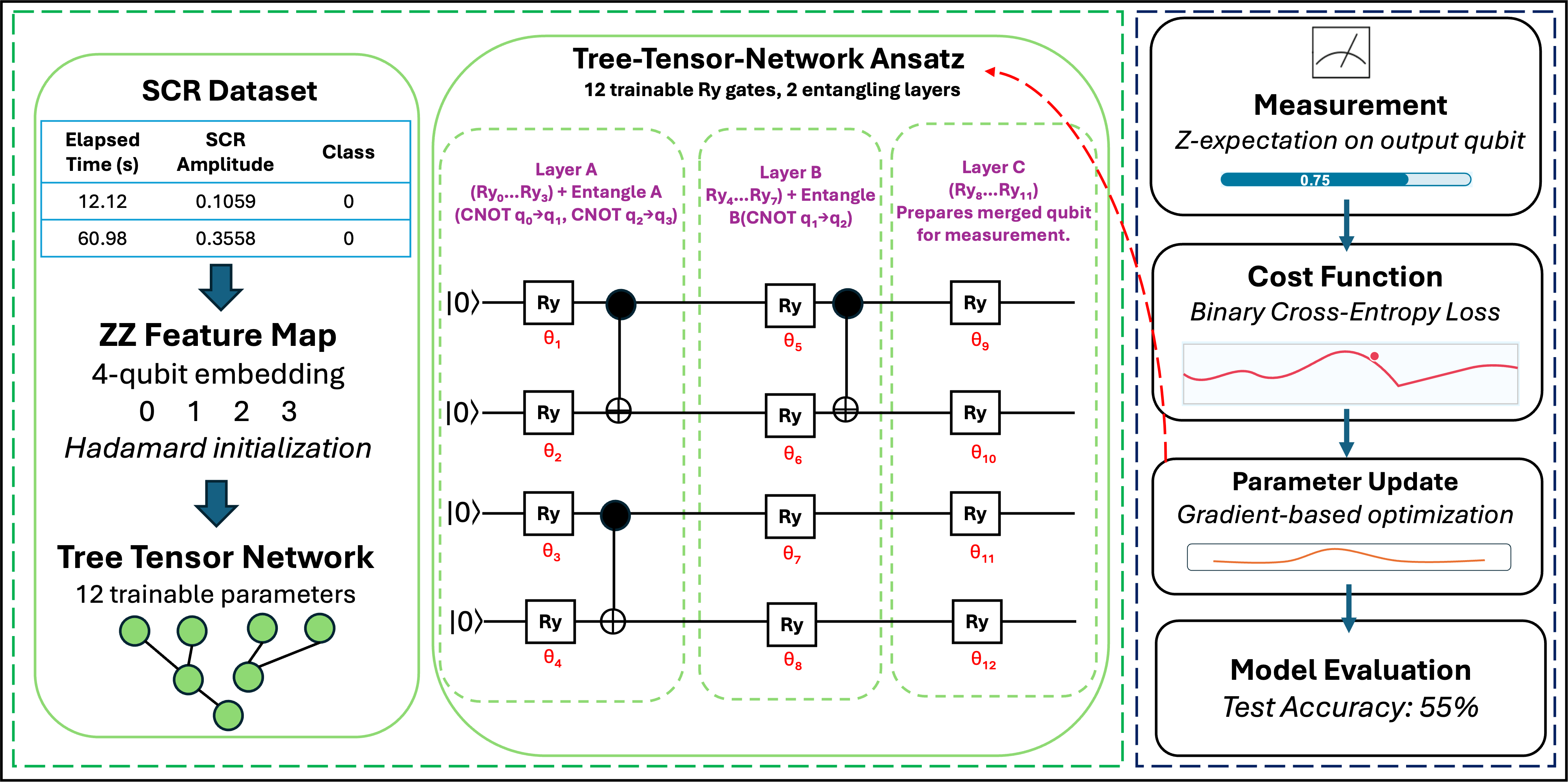}
    \caption{Quantum Neural Network}
    \label{fmb}
  \end{subfigure}

  \caption[Example workflow of quantum models with four qubits] 
  {%
    \textbf{Example workflow of quantum models with four qubits}\par   
    
  }
  \label{fig:workflow}
\end{figure*}

The two quantum machine learning models were built upon their classic version, i.e., a support vector machine with RBF and linear kernels, and a feed-forward neural network (see Fig. \ref{fig:workflow}). For the quantum versions of these classical models, we used a quantum kernel derived from three different feature mapping circuits, i.e., Amplitude-Encoding, Angle-Encoding and ZZ feature map. Additionally, for QNN we also used a variational quantum circuit with a Tree Tensor Network structure.

    {\color{black}%
      Fig.\;2(a) shows the workflow for QSVM. The raw SCR data (elapsed time and amplitude) go through
      data processing where they are scaled to [0, $\pi/2$] and $L_{2}$-normalized, then embedded
      into a four-qubit circuit via the ZZ feature map. Each qubit is initialized with a Hadamard
      gate, $R_{z}$ rotations encode the two normalized features, and CNOT–$R_{z}$–CNOT sequences
      introduce pairwise ZZ interactions ($x_{0}$–$x_{1}$ and $x_{2}$–$x_{3}$). This produces the
      quantum state $\lvert\phi(x)\rangle$ whose overlap defines the kernel
      $k(x,y)=\lvert\langle\phi(x)\vert\phi(y)\rangle\rvert^{2}$. The resulting kernel matrix is
      passed to a classical SVM solver (with both linear and RBF kernels) for binary stress
      classification, yielding 78\% training accuracy and 45\% test accuracy on our VR-based GSR
      dataset.

      Fig.\;2(b) shows the QNN built on a Tree-Tensor-Network ansatz. Starting from the same
      ZZ-encoded input and Hadamard initialization, the circuit applies two entangling layers
      interleaved with 12 trainable $R_{y}$ rotations arranged in a binary-tree structure. After the
      variational layers, the Pauli-$Z$ expectation on the designated output qubit is measured and
      mapped via a sigmoid to produce a stress-level probability. Parameters are optimized using
      binary cross-entropy loss and gradient-based updates, resulting in 55\% test accuracy and
      demonstrating the network’s ability to capture nonlinear correlations among SCR features.%
    }%

\subsection{Classical SVM (RBF and Linear)}
For the classic SVM approach, two different kernel functions were used, including, a radial basis function (RBF) kernel, used for nonlinear classification, and a linear kernel, used as a baseline linear classifier \cite{cortes1995}.

\emph{Feature Preprocessing and Training Process:} 
Min-Max scaler is used to convert all features into the range $[0, 1]$ to make the model comparable with the features and to avoid bias that could result from different ranges of different features. A random sample of 100 data points was taken from the dataset and split with an 80/20 (train/test) split to allow a balanced class distribution in the test set.

After a sensitivity analysis, both SVM models used a regularization parameter $C = 0.5$, and class weights were balanced to deal with any class imbalance. 
The RBF SVM kernel used $\gamma = 2.0$ (Gaussian kernel width),  where $\gamma$ controls the kernel’s spread or sensitivity, while the linear SVM uses a simple linear hyperplane in the original feature without the need for a $\gamma$ parameter \cite{grant2018}.

\subsection{Kernel Trick: Classical vs Quantum}
The kernel trick is a technique in machine learning where a nonlinear mapping to a high-dimensional space is used implicitly to enable linear separation of data that is not linearly separable in the original space \cite{grant2018}.

\subsubsection{Classical SVM and the RBF Kernel}
The RBF kernel implicitly maps inputs to a higher-dimensional feature space by considering all powers of the input \cite{hofmann2008kernel}
\begin{equation}
K(x,y) = \exp(-\gamma \|x - y\|^2).
\end{equation}

\subsubsection{Quantum SVM (QSVM) and the Quantum Kernel Trick}
The QSVM defines a quantum kernel
\begin{equation}
k(x,y) = |\langle\phi(x)|\phi(y)\rangle|^2
\end{equation}
where $|\phi(x)\rangle$ is the quantum feature state. The important advancement here is that the space of quantum states can be exponentially large, i.e., an $n$-qubit state is in a $2^n$-dimensional Hilbert space. This means that the quantum feature map can encode $n$ features into quantum state that resides in a  $2^n$  Hilbert space.

\subsection{Quantum Feature Maps and Data Encodings}
A key step in quantum machine learning is encoding classical data into quantum states. A quantum feature map is a mapping $x \mapsto |\phi(x)\rangle$ that transforms an input vector $x$ into a quantum state $|\phi(x)\rangle$ \cite{schuld2019}. This is implemented by a quantum circuit $|\phi_{\text{out}}\rangle = U |\phi_{\text{in}}\rangle$ (where $U$ represents the unitary operations forming the circuit) that depends only on the data. In our case we analyzed three different feature mapping quantum circuits.

\subsubsection{Angle Encoding}
For each feature $x_j$ in an input vector of length $n$, we allocate one qubit and rotate it by an angle proportional to $x_j$~\cite{schuld2020}. For example, applying a Y-rotation $R_y(x_j)$ on qubit $j$ (initially in $\ket{0}$) prepares
\begin{equation}
R_y(x_j)\ket{0}
= \cos\!\Bigl(\tfrac{x_j}{2}\Bigr)\ket{0}
+ \sin\!\Bigl(\tfrac{x_j}{2}\Bigr)\ket{1}
\end{equation}
Here, $\cos(x_j/2)$ and $\sin(x_j/2)$ are the probability amplitudes of $\ket{0}$ and $\ket{1}$, respectively. Since measurement probabilities are the squares of these amplitudes, they satisfy the normalization condition
\begin{equation}
\cos^2\!\Bigl(\tfrac{x_j}{2}\Bigr)
+ \sin^2\!\Bigl(\tfrac{x_j}{2}\Bigr)
= 1.
\end{equation}

This follows directly from the $R_y$ gate definition, where the rotation angle is halved inside the trigonometric functions. In this process, each classical feature value $x_j$ is mapped to a quantum rotation angle, embedding the data into the qubit amplitudes.

\subsubsection{Amplitude Encoding}
It embeds the entire classical feature vector into the amplitudes of a quantum state~\cite{rebentrost2014}.  
For an input vector $x = (x_0, x_1, \ldots, x_{2^n - 1})$ of length $2^n$, amplitude encoding prepares an $n$-qubit quantum state given by:
\begin{equation}
\ket{\phi(x)}
= \frac{1}{\sqrt{\sum_{k=0}^{2^n-1} \lvert x_k\rvert^2}}
  \sum_{k=0}^{2^n-1} x_k \,\ket{k}.
\end{equation}

Here:
\begin{itemize}
  \item \(x_k\) is the \(k\)th entry of the classical vector \(x\),
  \item \(\ket{k}\) is the \(n\)-qubit computational-basis state labelled by the integer \(k\),
  \item The prefactor \(\bigl(\sum_{k=0}^{2^n-1}\lvert x_k\rvert^2\bigr)^{-1/2}\) normalizes \(\ket{\phi(x)}\) so that its measurement probabilities sum to 1.
\end{itemize}

Because an \(n\)-qubit state must have unit norm, only \(2^n - 1\) of its real amplitudes can be chosen independently. This method uses \(n\) qubits to represent \(2^n\) features, allowing us to load large amounts of classical information into a small number of qubits for efficient data loading in our quantum algorithms.

\subsubsection{ZZ Feature Map}
It combines single-qubit and two-qubit gates \cite{Havlicek2019}:
\begin{itemize}
\item \textbf{Initialization:} Apply a Hadamard gate to each qubit (putting all into superposition).
\item \textbf{Feature encoding:} For each feature $x_j$ (normalized to an angle), apply $R_z(x_j)$ to qubit $j$.
\item \textbf{Entangling layer:} For every qubit pair $(j, k)$, apply a CNOT (control $j$, target $k$), then apply $R_z(x_j \cdot x_k)$ on the target qubit.
\end{itemize}

We can perform this process across several layers, alternating between rotation and entanglement,  to get our deeper future map that can capture the increase in the complex relationships amongst features from data. \cite{deBeaudrap2019pauli}.

\subsection{Quantum SVM (QSVM)}
Like classic SVM, the QSVM is also a kernel-based classifier where a quantum kernel is computed using a quantum feature map. Input data $x$ (with $n$ features) is mapped to a quantum state $|\phi(x)\rangle$ using the parameterized quantum circuit (feature map); the kernel is defined as $k(x,y) = |\langle\phi(x)|\phi(y)\rangle|^2$ where the kernel is the squared absolute value of the inner product between two quantum states, which means it measures how similar the states are by measuring how much they overlap. In our implementation, we encode four selected features into an eight-qubit system. 

\emph{QSVM Training Procedure:} 
After computing the quantum kernel matrix for all training data points, a classical SVM solver was used to find the optimal hyperplane in the quantum-induced feature space. To keep the fairness, we train the QSVM on the same samples as classical SVM (80 training samples, 20 test samples). Preprocessing is applied to optimize performance: Features were scaled to $[0,\pi/2]$ for rotation angles to ensure appropriate parameter ranges for quantum gate rotations, and an entanglement scaling factor (0.7) was applied in the ZZ map to avoid over complexity, following this, each feature vector was normalized using $L_2$ normalization to match the quantum state requirements. 

\subsection{Classical Neural Network}
The classical neural network model used in this study is a supervised deep learning approach based on a standard feed-forward architecture, implemented using TensorFlow/Keras.

All four features were taken as input data for the neural network, and for this implementation, the amplitude classes were reduced into two groups: low and high amplitude responses to simplify the problem and ensure better model performance. The target variable was defined for a binary classification task, coded as \texttt{amp\_class} (low vs. high). The resulting dataset was balanced.

\emph{Network Architecture:} 
After a sensitivity analysis, the final neural network architecture consisted of three fully connected layers (excluding the input layer):
\begin{itemize}
    \item \textbf{Input layer:} Accepts 4 feature values.
    \item \textbf{Hidden layer 1:} 12 neurons with ReLU activation.
    \item \textbf{Dropout layer:} 30\% dropout after the first hidden layer.
    \item \textbf{Hidden layer 2:} 6 neurons with ReLU activation.
    \item \textbf{Dropout layer:} 30\% dropout after the second hidden layer.
    \item \textbf{Output layer:} Single neuron with sigmoid activation.
\end{itemize}


\paragraph{Training Procedure:}
The network was trained using the Adam optimizer with several regularization techniques:
\begin{itemize}
    \item \textbf{Early stopping:} Training was monitored for improvements in validation loss.
    \item \textbf{Learning rate scheduling:} When validation loss plateaued, the learning rate was reduced.
    \item \textbf{L2 weight decay:} Regularization was applied to penalize large weight values.
\end{itemize}

\subsection{Quantum Neural Network (QNN) with \texttt{ZZFeatureMap} and Tree Tensor Network}
The Quantum Neural Network model combines the quantum encoding (same ZZ FeatureMap on eight-qubits) with a variational quantum circuit (a Tree Tensor Network, TTN, ansatz).

\emph{Tree Tensor Network Ansatz:} 
The variational circuit employed was the \textit{Tree Tensor Network} (TTN) ansatz, a structured, layered circuit inspired by tensor networks. In a TTN, qubits are processed in a hierarchical binary tree pattern \cite{cheng2019}. Our implementation uses two layers of trainable single-qubit Y-rotations (i.e., $R_y$ gates), separated by an entangling layer of CNOT gates:

\begin{itemize}
    \item \textbf{First variational layer:} Applies $R_y(\theta_j)$ gates on each of the eight qubits, using two parameters per qubit.
    \item \textbf{Entangling layer:} Use CNOT gates to entangle pairs of qubits in a tree-like pattern.
    \item \textbf{Second variational layer:} applies another set of eight $R_y$ rotations with  new parameters.
\end{itemize}

This gives us 24 trainable parameters. This model was selected because it is well-suited for binary classification tasks similar to our analysis of the low versus high stress classes,  and encodes an  inductive bias where feature interactions are combined in a hierarchical  structure, allowing the model to capture layered relationships between features  \cite{cheng2019}.

\emph{Training the QNN:} 
We trained the QNN
(learning rate 0.05) using full-batch gradient updates, where full training set is used when computing each update. 
After the variational layers, the expectation value of the Pauli-Z operator was measured on a designated output qubit. This Z value is then mapped to the range $[0,1]$ and interpreted as the probability of being classified in a high-amplitude class. A threshold of 0.5 on this probability was applied to assign the final class label (values above 0.5 were classified as high amplitude, and those below 0.5 as low amplitude) \cite{farhi2018}.

\begin{figure*}
\centering
 \begin{subfigure}[b]{0.3\textwidth}
 \centering
    \includegraphics[width=\linewidth]{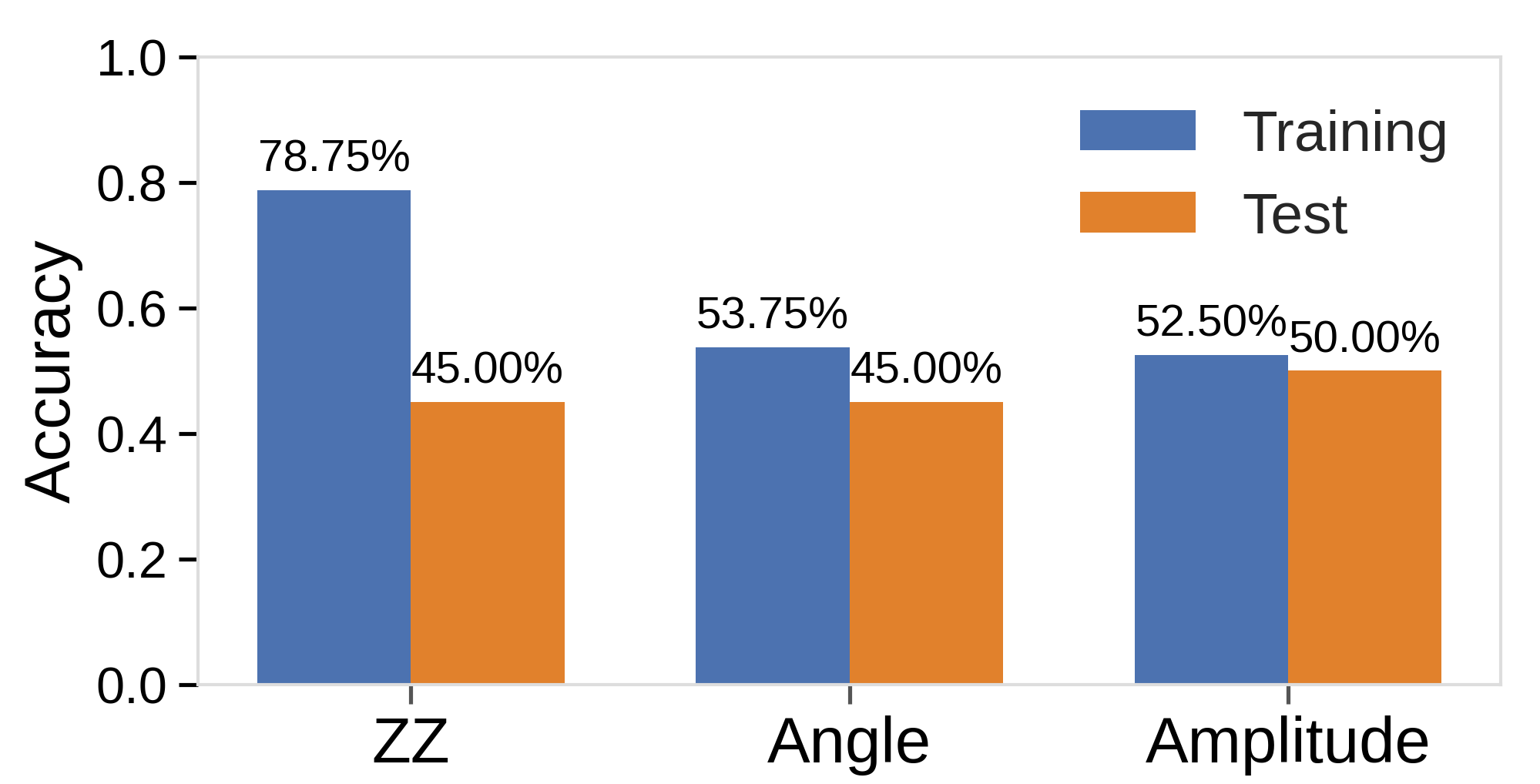}
    \caption{Model Accuracy}
    \label{fmaa}
    \end{subfigure}
    \hfill
     \begin{subfigure}[b]{0.3\textwidth}
 \centering
    \includegraphics[width=\linewidth]{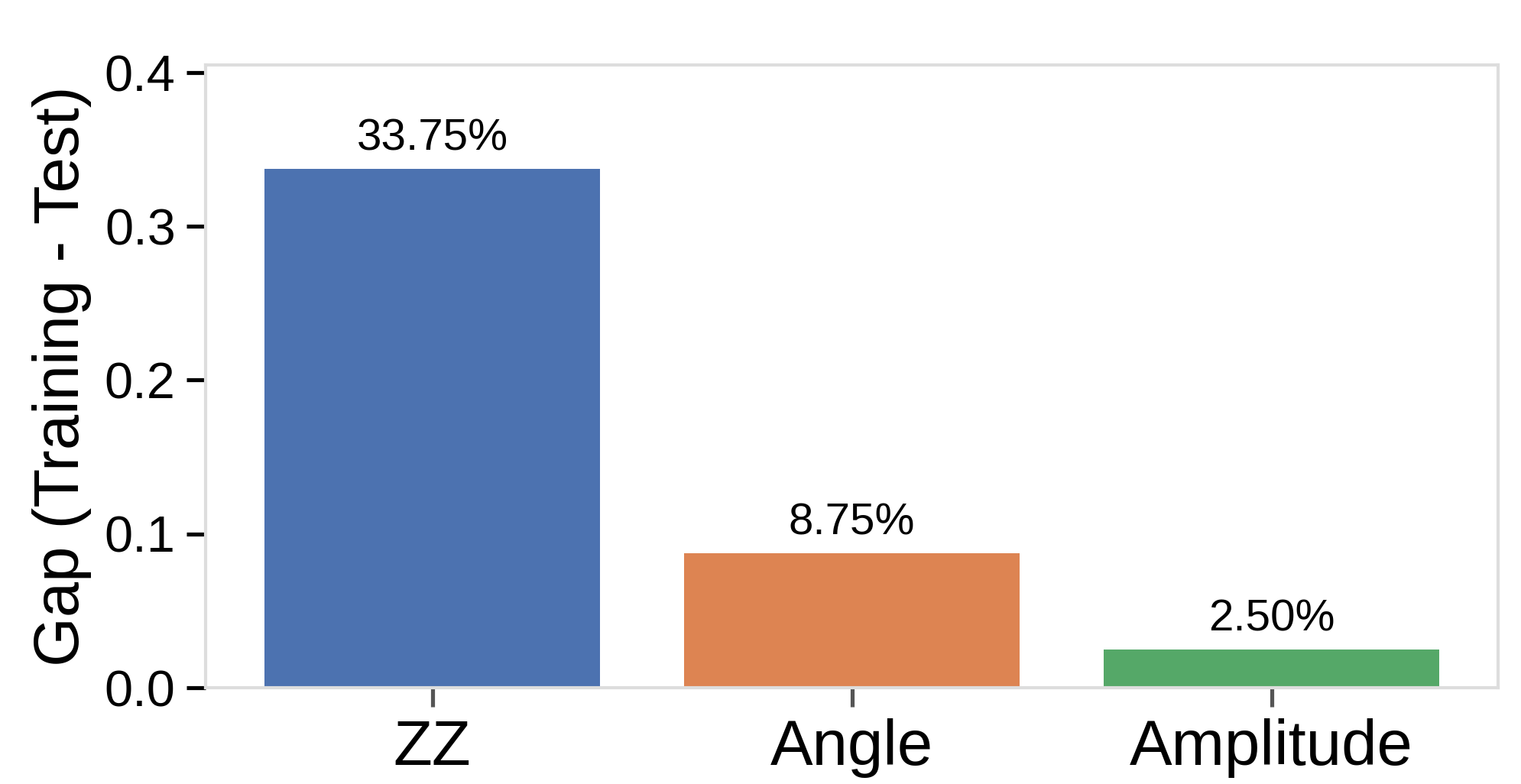}
    \caption{Generalization Gap}
    \label{fmbb}
    \end{subfigure} 
\hfill
 \begin{subfigure}[b]{0.3\textwidth}
 \centering
    \includegraphics[width=\linewidth]{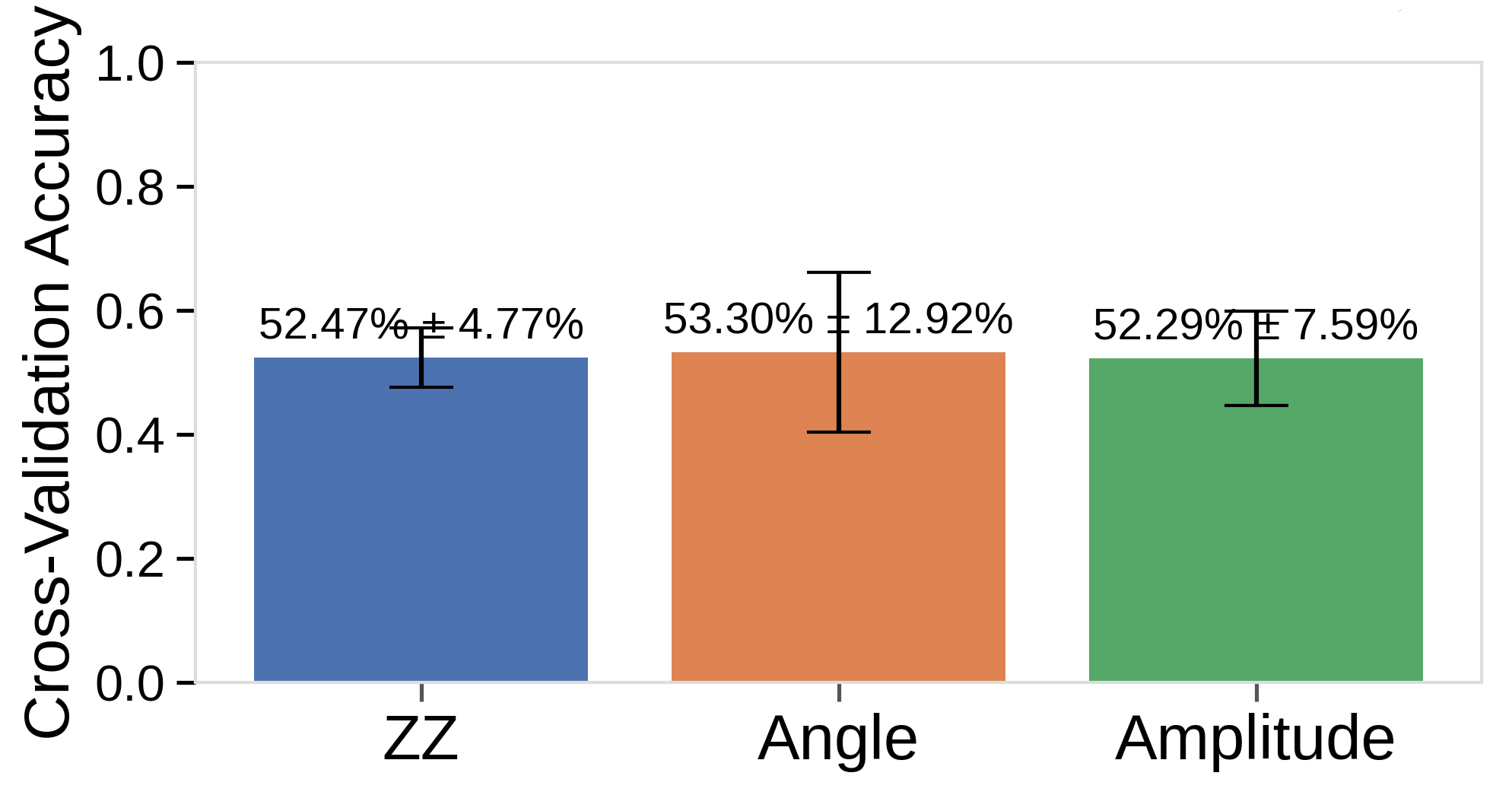}
    \caption{Cross validation accuracy}
    \label{fmc}
    \end{subfigure}
\caption{Performance comparison of quantum feature maps}
\label{fig:featuremaps}
\end{figure*}
\begin{figure*}
    \centering
 \begin{subfigure}[b]{0.24\textwidth}
 \centering
    \includegraphics[width=\linewidth]{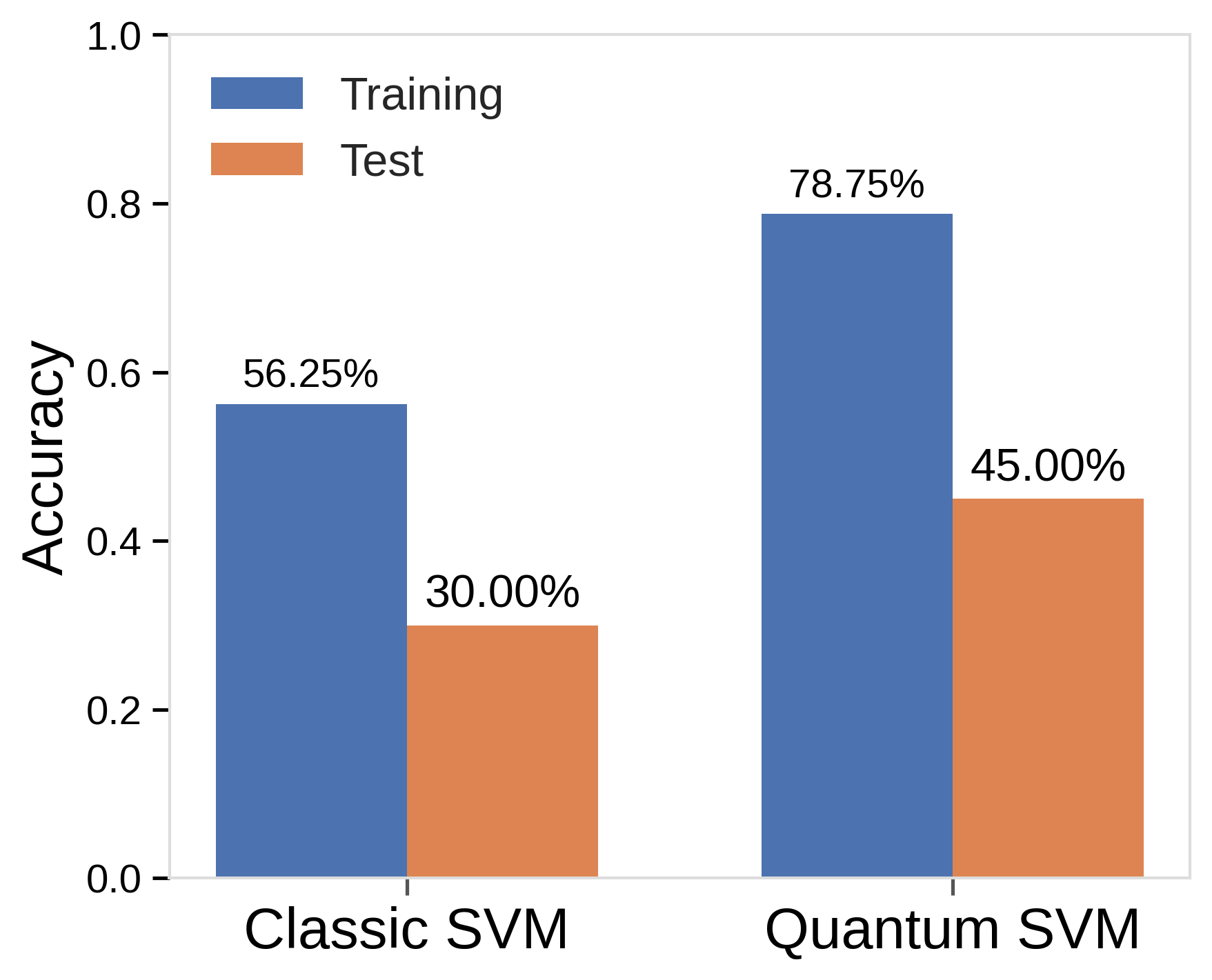}
    \caption{Model Accuracy}
    \label{svma}
    \end{subfigure}
     \begin{subfigure}[b]{0.24\textwidth}
 \centering
    \includegraphics[width=\linewidth]{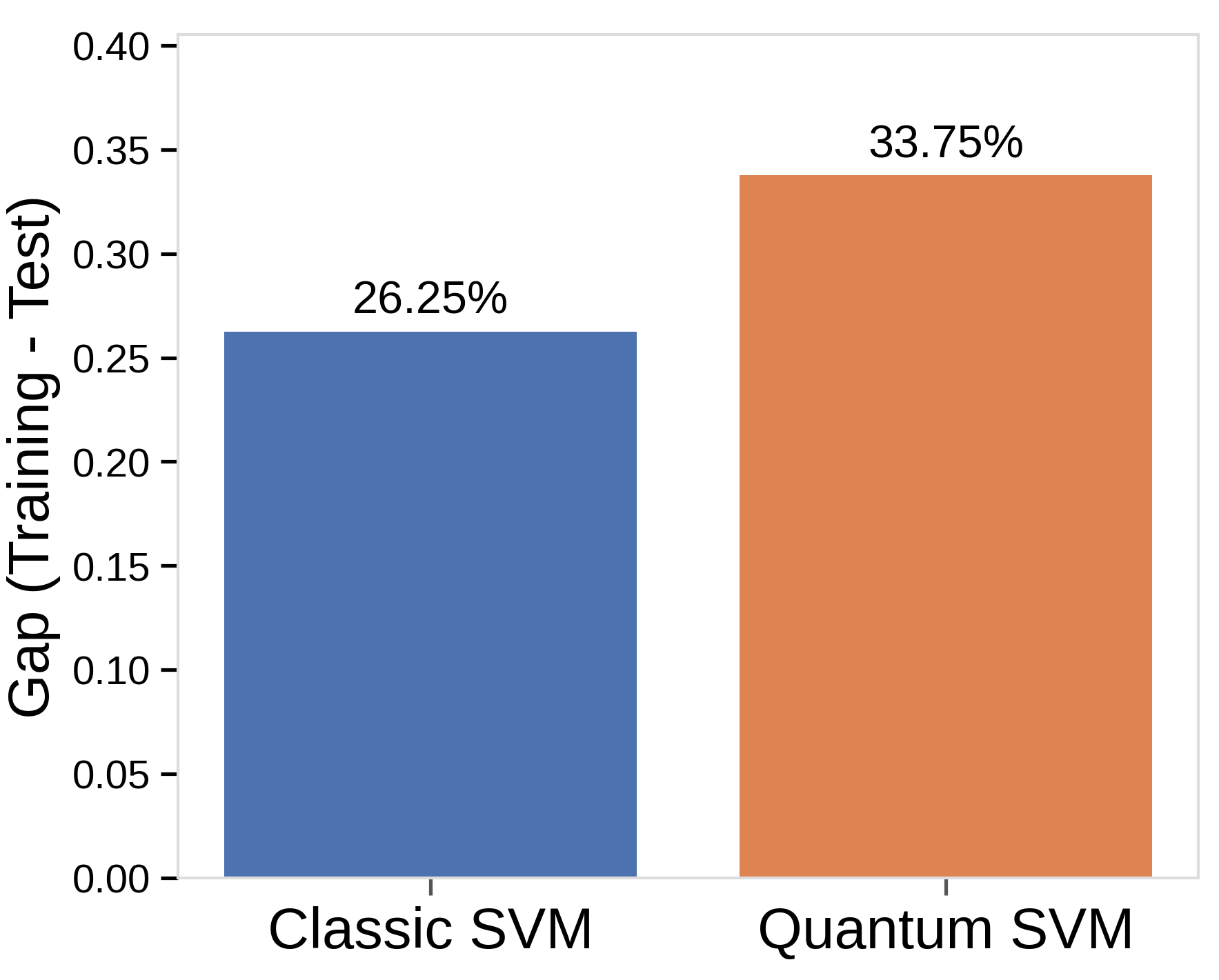}
    \caption{Generalization Gap}
    \label{svmb}
    \end{subfigure}    
     \begin{subfigure}[b]{0.24\textwidth}
 \centering
    \includegraphics[width=\linewidth]{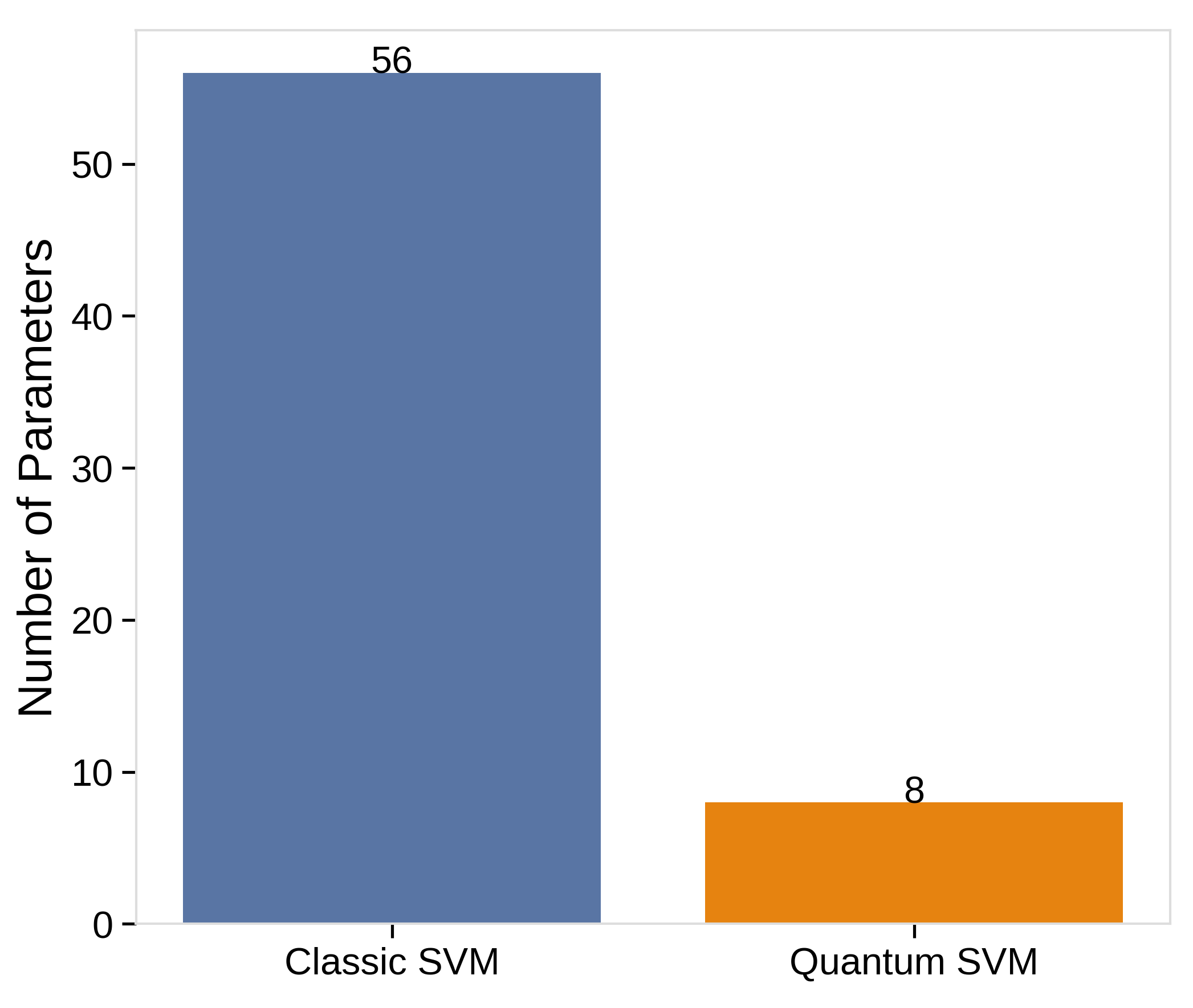}
    \caption{Model Complexity}
    \label{svmc}
    \end{subfigure}    
     \begin{subfigure}[b]{0.24\textwidth}
 \centering
    \includegraphics[width=\linewidth]{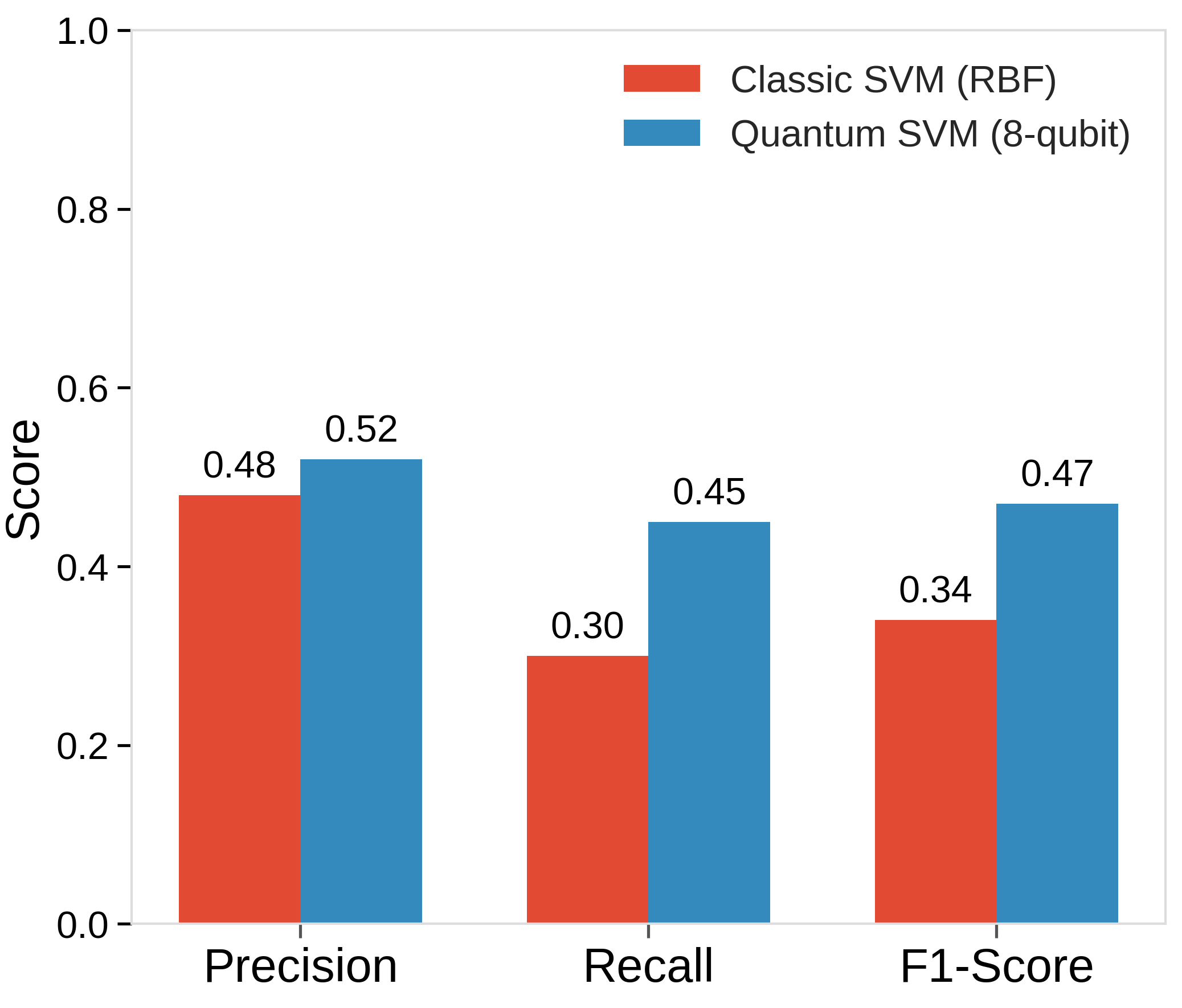}
    \caption{Precision, Recall \& F1}
    \label{svmd}
    \end{subfigure}    
\caption{Performance comparison of classical SVM and QSVM models.}
\label{fig:svm vs qsvm}
\end{figure*}
\begin{figure*}
\centering
\begin{subfigure}[b]{0.24\textwidth} 
\includegraphics[width=\textwidth]{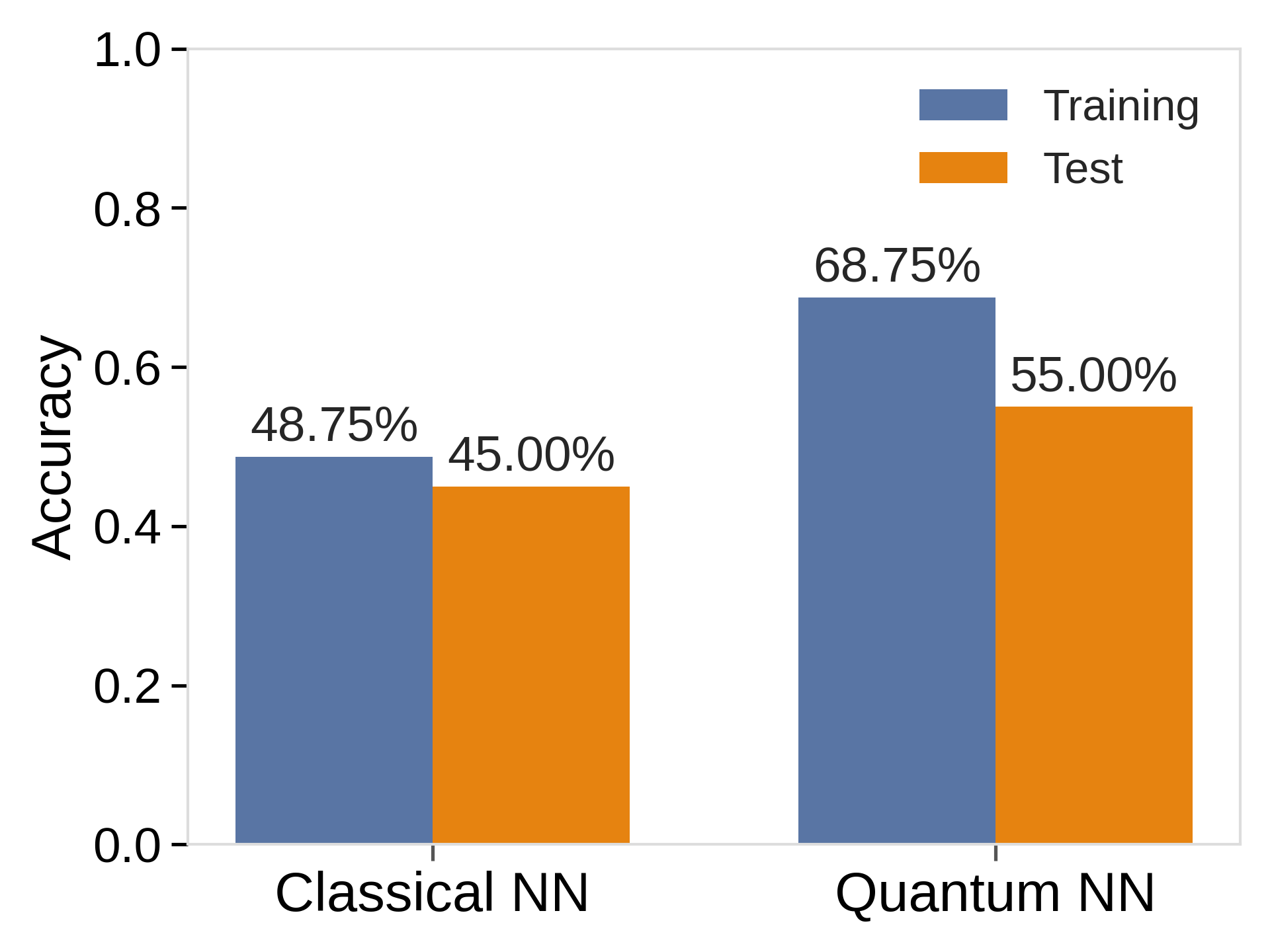}
\caption{Model Accuracy}
\label{fig:qna}
\end{subfigure}
\hfill
\begin{subfigure}[b]{0.24\textwidth} \includegraphics[width=\textwidth]{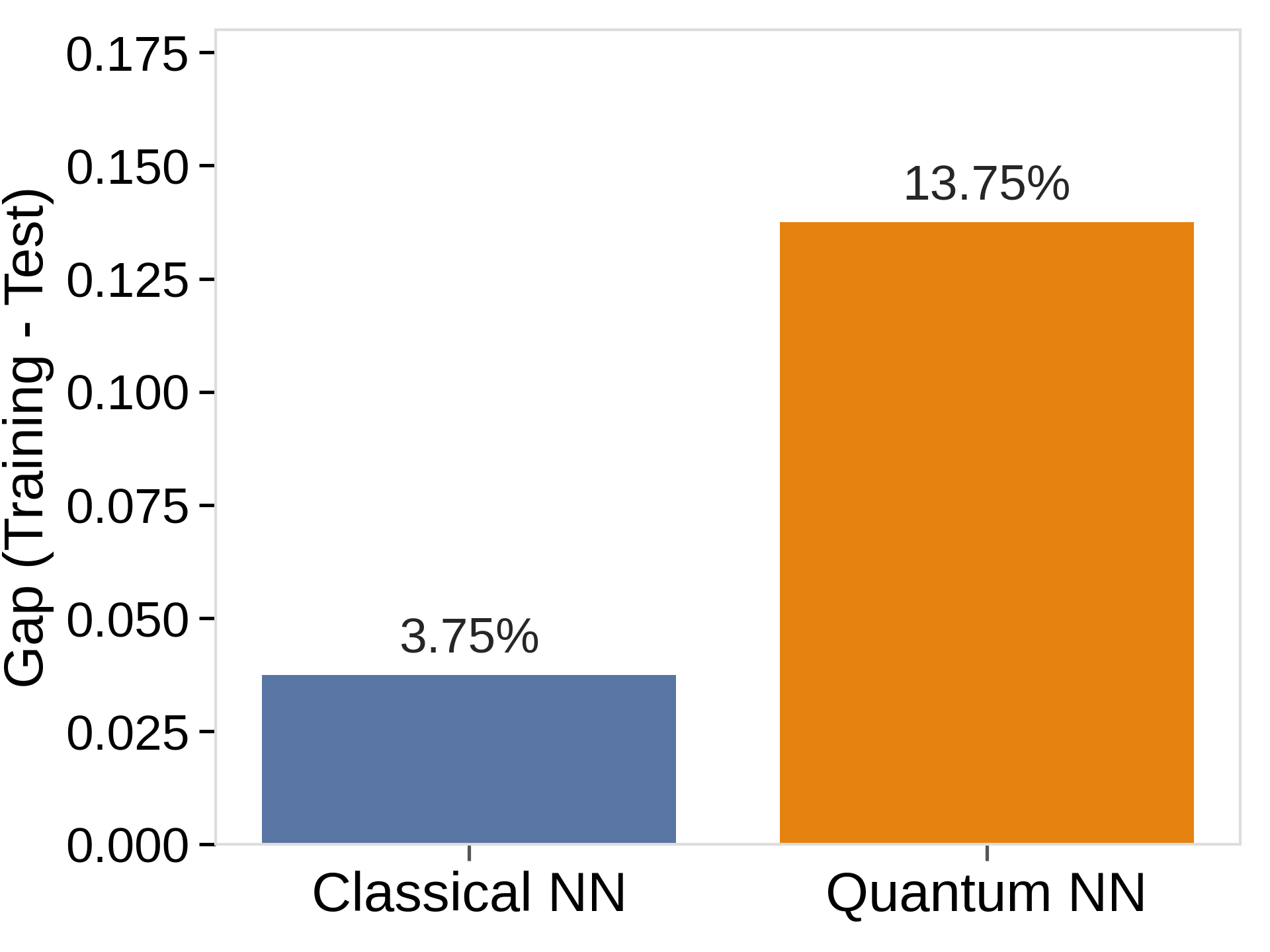}
\caption{Generalization Gap}
\label{fig:qnb}
\end{subfigure}
\hfill
\begin{subfigure}[b]{0.24\textwidth} 
\includegraphics[width=\textwidth]{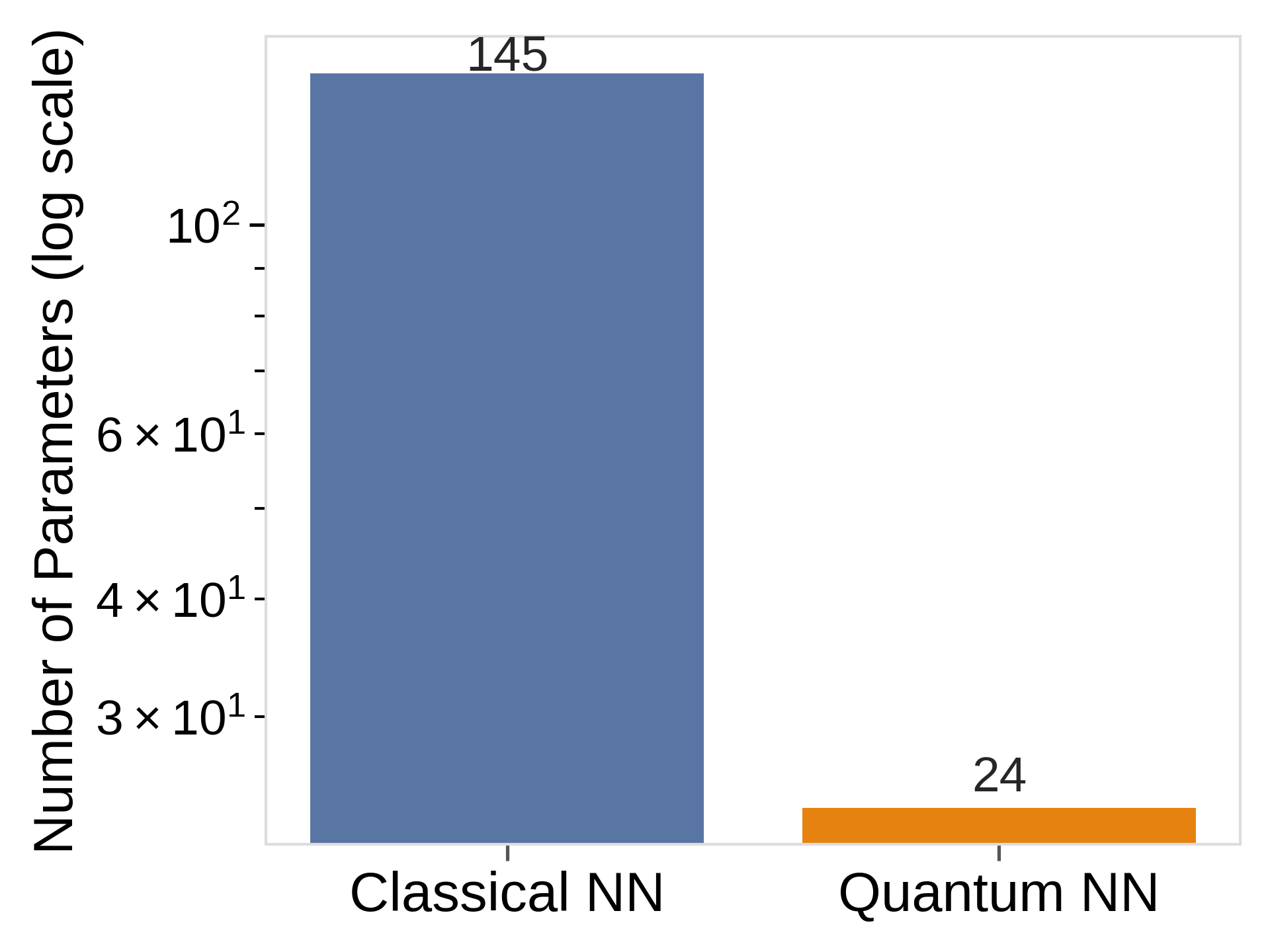}
\caption{Model Complexity}
\label{fig:qnc}
\end{subfigure}
\hfill
\begin{subfigure}[b]{0.24\textwidth} 
\includegraphics[width=\textwidth]{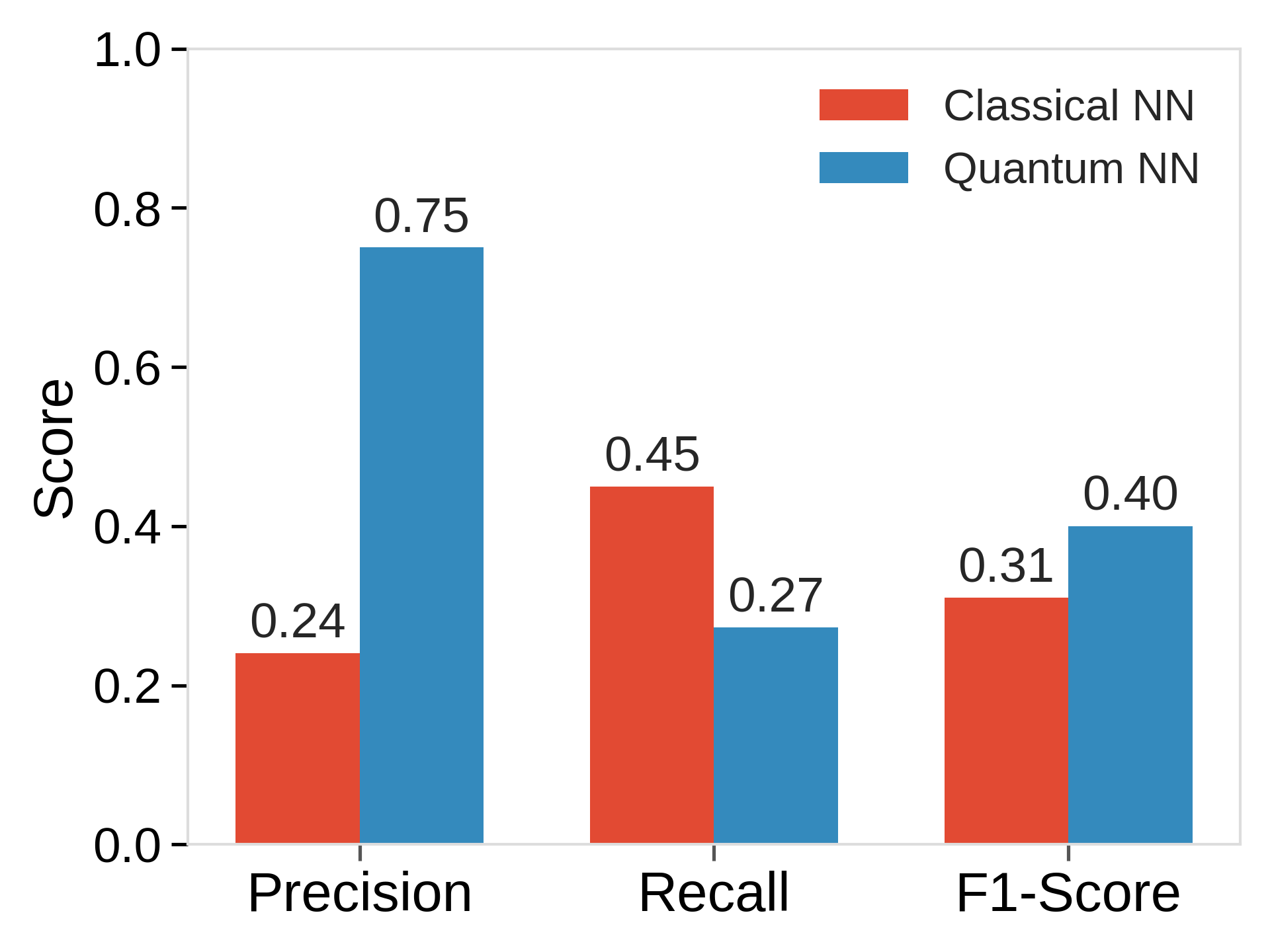}
\caption{Precision, Recall \& F1}
\label{fig:qnd}
\end{subfigure}
\caption{Performance comparison of classical NN and QNN models.}
\label{fig:qnn_vs_nn}
\end{figure*}

\section{Model Performance Analysis}
Fig.~\ref{fig:featuremaps} reports the performance of the three feature maps used in the QSVM model. The ZZ map achieved the highest training accuracy (78.75\%), but suffered from a large generalization gap (33.75). In contrast, angle encoding (8.75\% gap) and amplitude encoding (2.50\% gap) exhibit smaller gaps and superior test performance, underscoring the risk of high‐complexity mappings on small datasets and the value of restrained encodings. We decided to use ZZ map in the rest of the analysis, as it can capture pair-wise interactions, uses qubits entanglement, has medium circuit complexity, and has the highest training accuracy compared to other feature mapping quantum circuits.

After completing the training and evaluation of all the models, we compare their performance on the SCR dataset classification task. The goal is to look into the key metrics of accuracy, generalizability, precision, recall, and F1-score to assess not only the overall correctness of the models, but also the quality of positive vs. negative predictions. Fig.~\ref{fig:svm vs qsvm} and \ref{fig:qnn_vs_nn} provide the detailed visualization of the results.  


The classical SVM with RBF kernel achieved a test accuracy of 30\%, while its quantum version, the QSVM achieved 45\%. The QNN achieved the highest test accuracy at 55\%, while its classic version achieved 45\%. In g        
                         eneral, it can be observed that both quantum models achieved a higher accuracy when compared to their classic counterparts. In terms of the training accuracy, QSVM achieved the highest accuracy (78.75\%), but suffered from overfitting, which is evident from its lower test accuracy. 

The classical neural network demonstrated strong performance in both precision and recall, achieving an overall precision/recall of 0.42/0.24 and an F1-score of 0.31. 
The QSVM with 45\% accuracy had very low recall 0.45, showing poor detection of either the high or low amplitude classes. For the QNN which performed binary classification with 55\% accuracy on a balanced set, its precision and recall for the ``high amplitude" class were both around 0.75.

Looking into the generalization gaps in the analysis between training and test accuracy, a larger generalization gap was observed in the QSVM with the ZZ feature map (33.75\% gap) compared to the classical linear SVM (6.25\% gap). The neural network models experience lower differences in gap size, with QNN having  13.75\% gap in comparison to the classic NN (3.75\% gap). Despite the QSVM having higher training accuracy (78.75\%) than the classical SVMs, this advantage didn't translate to test performance, highlighting its poor generalization.

Quantum models appeared to overfit more than classical models. The QSVM had a high overfiting, with training accuracy at 78.75\% and test accuracy at 45\%. The QNN showed a significant but smaller gap 14\% (69\% vs.\ 55\%). The classical models, however, especially the neural network, had very small overfitting, with only a 3.75\% gap between training and testing accuracy.
The classical models demonstrated better generalization than quantum models. However, both QNN and QSVN required significantly fewer parameters than their classic counterpart, showing a much higher efficiency in terms of information usage.
Both classical SVM and QSVM used four features, but QSVM also used a more complex kernel and showed significant overfitting. This complexity, combined with the fixed quantum kernel and lack of regularization, likely contributed to the QSVM's poor generalization. For the neural networks, both the classical NN and QNN utilized all four features. One positive aspect is the parameter efficiency of the QNN, as it used only 24 parameters to achieve 55\% accuracy, while the classical NN needed 145 parameters to reach 45\% suggesting QNNs can model complex patterns with fewer resources.

With higher test accuracy (55\% for QNN vs. 45\% for QSVN), similar percentage reduction in parameters (83\% for QNN and 85\% for QSVN), and smaller generalization gap (14\% for QNN vs 3.75\% and QSVN), the quantum variational approach is more effective in modelling the complexity of our dataset compared to the fixed quantum kernel approach. A possible reason for this is that the QNN could adapt its decision boundary during training, whereas the QSVM's kernel was fixed and possibly not well-suited for the task (or overfitted due to lack of regularization). This brings the need for a further discussion in quantum machine learning to see whether fixed quantum kernels or variational circuits are better for learning, with the right choice often depending on the structure and complexity of the data. 

\section{Conclusion}
The Quantum SVM achieved 78.75\% training accuracy but only 45\% test accuracy, showing the large generalization gap due to overfitting. The Quantum Neural Network outperformed the QSVM, with 55\% test accuracy and showed better generalizability.
From a theoretical perspective, the fixed complexity of the QSVM's kernel limited its adaptability to outperform a classical RBF kernel. Quantum feature maps can embed data into higher-dimensional spaces that can be utilized to enhance the analysis capability, but if they are not regularized properly, overfitting issues may occur. Both quantum models used fewer parameters than their classic counterparts and showed better test accuracy. 

All quantum experiments were conducted on Pennylane quantum emulator running on a classical computer. Due to the scale of underlying quantum mechanics, we were thus limited to a maximum eight-qubits circuit design only. This likely affected the QSVM and QNN, as larger feature maps and more qubits could not be explored. The emulator's low speed also restricted hyperparameter tuning and repeated runs. Running our models on real quantum hardware, which continues to scale up in terms of qubit counts, despite current noise related issues, could allow for achieving higher model accuracy. Being able to access to 20–30 high-quality qubits may allow for more expressive kernels and variational circuits, achieving a much higher performance. Early demonstrations on real quantum processors have already shown potential for quantum machine learning to outperform classical methods as hardware continues to advance \cite{nature2019quantum}. Furthermore, better designs and regularization strategies continue to improve the overall performance of classical models. Future research will look into improving quantum feature maps and circuit design using these regularizations, while investigating other intelligent transportation systems problems where quantum approaches can provide advantages. 
\textcolor{black}{The classical–quantum comparison can be further extended to other quantum-circuit-compatible methods, such as quantum k-nearest neighbours for classification and exploration of quantum regression models for continuous stress prediction. We also recommend, testing how well the models scale to larger VR-based GSR datasets and how stable they are in noisy intermediate-scale quantum (NISQ) era, using basic error-mitigation techniques.}

\bibliographystyle{ieeetr}
\bibliography{bibliography}

\end{document}